\documentclass[10pt,twocolumn,letterpaper]{article}

\usepackage{cvm}
\usepackage{times}
\usepackage{epsfig}
\usepackage{graphicx}
\usepackage{amsmath}
\usepackage{amssymb}
\usepackage{siunitx}
\usepackage{capt-of}
\usepackage{placeins}
\usepackage{wrapfig}

\usepackage[pagebackref=true,breaklinks=true,letterpaper=true,colorlinks,bookmarks=false]{hyperref}

\newcommand{\keywords}[1]{{\bf \emph{Keywords: #1}}}

\cvmfinalcopy %

\def\cvmPaperID{181} %

\ifcvmfinal\fi

\newif\ifconference
\conferencetrue 

\begin{document}

\title{ExCellGen: Fast, Controllable, Photorealistic 3D Scene Generation from a Single Real-World Exemplar}

\author{Cl\'ement Jambon\footnotemark[1]\\
Massachusetts Institute of Technology\\
Cambridge, USA\\
{\tt\small cjambon@mit.edu}
\and
Changwoon Choi\\
Seoul National University\\
Seoul, South Korea\\
{\tt\small changwoon.choi00@gmail.com}
\and
Dongsu Zhang\\
Seoul National University\\
Seoul, South Korea\\
{\tt\small 96lives@gmail.com}
\and
Olga Sorkine-Hornung \\
ETH Zurich \\
Zurich, Switzerland \\
{\tt\small sorkine@inf.ethz.ch}
\and 
Young Min Kim\\
Seoul National University\\
Seoul, South Korea\\
{\tt\small youngmin.kim@snu.ac.kr}
}

\newcommand{\methodname}{ExCellGen}

\twocolumn[{%
\renewcommand\twocolumn[1][]{#1}%
\maketitle
\begin{center}
  \includegraphics[width=\textwidth]{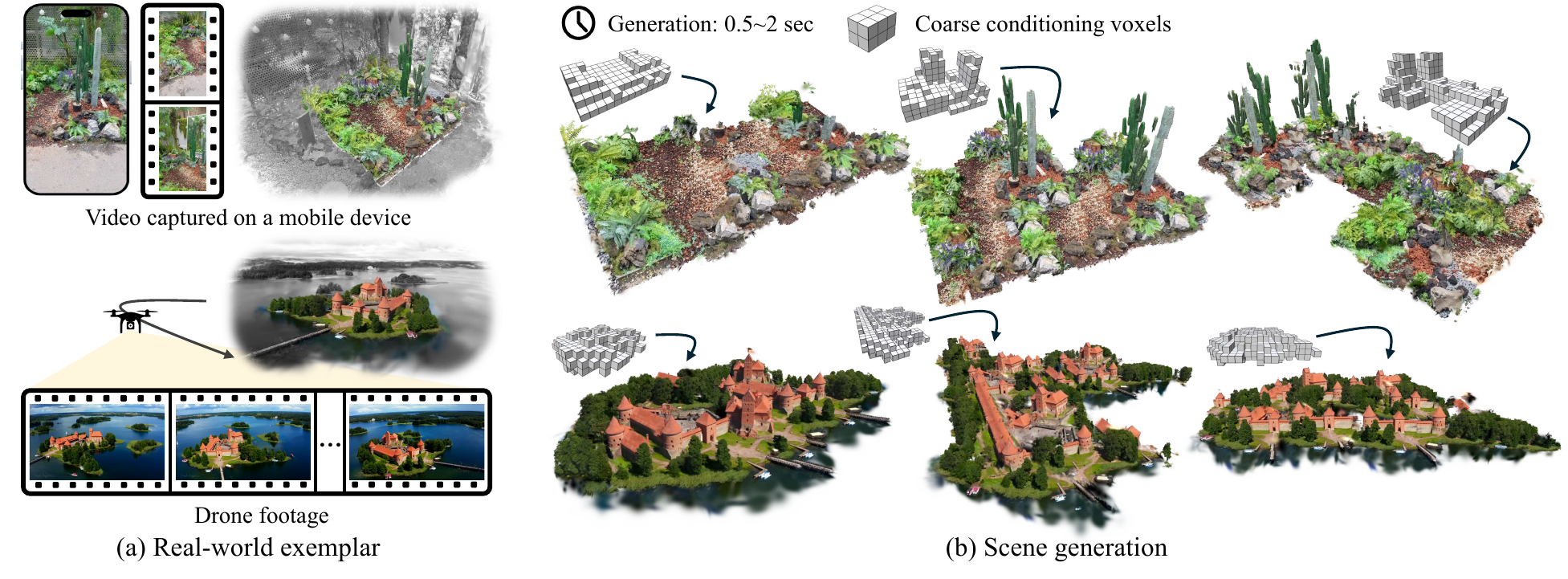}
  \captionof{figure}{
    We propose \methodname, a framework for fast, controllable, and photorealistic 3D scene generation from real-world exemplars.
    We convert casual video inputs (e.g., drone or mobile footage) into a high-quality 3D representation using 3D Gaussian splats.
    Within our GUI editor, users can extract regions from which they wish to generate variations.
    Our two-stage amortized generation strategy first uses a per-scene Generative Cellular Automaton (GCA)~\cite{gca} to produce a sparse volume of featurized voxels, then composites the final appearance via sparse patch-based remapping from the original scene.
    Given coarse conditioning voxels, users can synthesize new variations in just seconds in a fully interactive way.
  }
  \label{fig:teaser}
\end{center}
}]

{
\renewcommand{\thefootnote}{*}
\footnotetext[1]{Work done at Seoul National University and ETH Zurich.}
}

\begin{abstract}

Photorealistic 3D scene generation is challenging due to the scarcity of large-scale, high-quality real-world 3D datasets and complex workflows requiring specialized expertise for manual modeling. 
These constraints often result in slow iteration cycles, where each modification demands substantial effort, ultimately stifling creativity.
We propose a fast, exemplar-driven framework for generating 3D scenes from a single casual input, such as handheld video or drone footage. 
Our method first leverages 3D Gaussian Splatting (3DGS) to robustly reconstruct input scenes with a high-quality 3D appearance model.
We then train a per-scene Generative Cellular Automaton (GCA) to produce a sparse volume of featurized voxels, effectively amortizing scene generation while enabling controllability. 
A subsequent patch-based remapping step composites the complete scene from the exemplar's initial 3D Gaussian splats, successfully recovering the appearance statistics of the input scene.
The entire pipeline can be trained in less than 10 minutes for each exemplar and generates scenes in 0.5-2 seconds.
Our method enables interactive creation with full user control, and we showcase complex 3D generation results from real-world exemplars within a self-contained interactive GUI.

\end{abstract}

\keywords{Interactive Method, Generative Cellular Automata, 3D Gaussian Splatting, Scene Generation, Patch-based Synthesis}

\def\EDIT#1{{#1}}

\newcommand{\gcastate}{X}
\newcommand{\voxelset}{\mathbf{V}}
\newcommand{\featureset}{\mathbf{Z}}
\newcommand{\probcond}{\,|\,}

\newcommand{\patchocc}{\mathbf{O}}
\newcommand{\patchfeat}{\mathbf{F}}

\section{Introduction}

Generating realistic 3D scenes is key to many applications in 3D content creation, visual effects, robotics~\cite{lucidsim}, and autonomous driving~\cite{nvidia-cosmos}.
Generative modeling has recently demonstrated remarkable progress in creating photorealistic images, largely owing to the tremendous data and development of deep learning network architectures.
In contrast, achieving realism in 3D scene generation faces several fundamental challenges.
Unlike images or photos, 3D models are limited in scale and quality.
Available 3D datasets are often synthetic, and composed of object-centric meshes with simple textured appearance~\cite{objaverse}. 
In contrast, existing real-world datasets are mostly confined to specific distributions, including indoor scenes~\cite{Matterport3D}, urban footage~\cite{kitti-360} or object-centric captures~\cite{co3d, uco3d, gco}.
This falls short of the diversity found in real-world scenes, rich in intricate geometry and natural elements like vegetation, which are notoriously difficult to capture with meshes or represent using conventional 3D assets.

In this work, we propose to formulate the task of 3D scene generation as an exemplar-based synthesis task, building on a rich line of research in texture synthesis~\cite{texsynth, texture-efros}. 
We target real-world captures in the form of casual video footage. 
All of our input exemplars are derived from footage recorded using smartphones or drones, demonstrating the practicality of our approach. 
To reconstruct 3D scenes, we build on the recent success of 3D Gaussian Splatting (3DGS)~\cite{3dgs}. 
Gaussian splats enable fast retrieval and rendering and lead to a photorealistic appearance. 
They accurately reflect the user’s intent by faithfully reproducing the images captured by the user, providing unambiguous control in selecting the ingredients of scene generation. 
Moreover, 3D Gaussians are explicit representations that can be directly manipulated, transported, and composited.

Previous approaches for exemplar-based synthesis demonstrate results on clean synthetic or pre-processed scenes, and typically support only mesh-like or SDF-based exemplars~\cite{shapeshifter, singleshapegen, sin3dm}.
Even when supporting radiance fields, methods such as Sin3DGen~\cite{sin3dgen} still require a mesh to guide generation.
This requirement limits applicability to real-world scenes, where appearance may be fuzzy in some regions. %
Moreover, casually captured scenes often contain complex backgrounds and surrounding objects with unclear segmentation boundaries.
We thus design intuitive means to select a desired region from the input exemplar scene by distilling and clustering semantic feature maps from large vision models~\cite{dino}.
This provides both control and additional guidance during generation.

Synthesizing 3D scenes is often memory and computationally intensive, as it requires inferring 3D geometry, leading to slow, iterative processes (e.g., 1-3 minutes for Sin3DGen).
We propose a fast and controllable two-stage generation strategy that enables interactive feedback within seconds. First, we generate a sparse voxel volume that captures the mesoscale structure of the scene, abstracting away fine-grained details and amortizing the generation task. Then, we reconstruct high-frequency appearance by compositing 3D Gaussians from the exemplar, guided by the generated voxels.

More precisely, in the first stage, we adapt Generative Cellular Automata~\cite{gca} (GCA) to controllable scene generation to efficiently generate a sparse 3D volume of voxels.
The generative kernels of GCA are trained from a single exemplar and are highly efficient in both training (less than 10 minutes) and generation (0.5-2 s).
This neural approach offers two additional advantages.
First, contrary to cascaded generation strategies~\cite{sin3dgen}, our generative network is naturally trained to reflect conditional signals, which we materialize as coarse voxels that describe the location and shape of the desired output scene.
Second, neural approaches generalize better across various distributions without the need for cumbersome parameter tuning.
In the second stage, we recover the final appearance of the scene by introducing a novel sparse and efficient patch-based consistency step, which results in smooth transitions between voxels. 
The final scene is converted into 3DGS with high-frequency details by transporting and compositing 3D Gaussians from the exemplar scene according to the generated voxels.
We demonstrate the practicality and capabilities of our method on a variety of real-world scenes captured using smartphones or drones, and manipulated directly within our interactive GUI editor.

In summary, our contributions are as follows:
(a)	We propose an efficient 3D scene generation technique in which exemplar scenes are directly captured from casual real-world videos;
(b)	Our method is controllable, inviting user controls in capturing, selecting, and compositing the scene;
(c)	We formulate a lightweight generative model from a single exemplar, allowing scene synthesis at interactive rates (approximately 0.5-2 s);
(d)	We provide a fully self-contained GUI to demonstrate the practicality of the method.
\EDIT{The source code is available at \href{https://github.com/clementjambon/excellgen}{https://github.com/clementjambon/excellgen}.}

\begin{figure*}
    \centering
    \includegraphics[width=\textwidth]{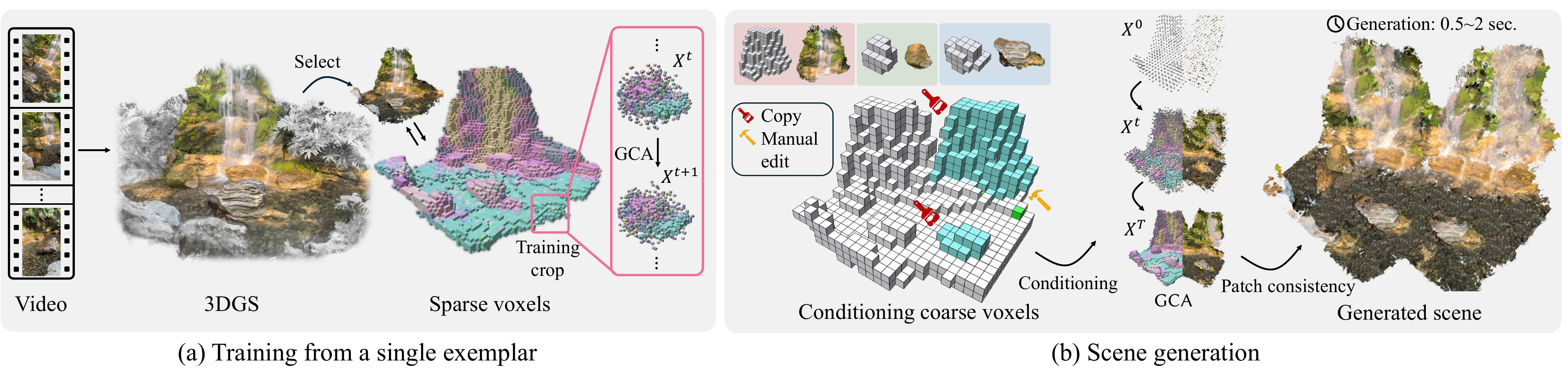}
    \caption{
        Overview of our pipeline. 
        \emph{Training phase.} (left) From a casual video, an exemplar scene is reconstructed as 3D Gaussians augmented with DINO features. 
        After selecting the region of interest within an interactive editor, the 3D Gaussians are converted into a sparse volume of voxels. A per-scene Generative Cellular Automaton (GCA) is then efficiently trained on random crops of the scenes in under 10 minutes.
        \emph{Generation phase.} (right)  A set of coarse conditioning voxels, either copied from parts of the exemplar or manually edited, is provided to the pre-trained GCA.
        The GCA generates a novel sparse volume of featurized voxels. 
    These voxels are then remapped to the exemplar’s 3D Gaussians using a sparse patch-based consistency step.
    }
    \label{fig:overview}
\end{figure*}

\section{Related Works}
\label{sec:related_works}

\paragraph{3D scene representations}
Generating high-quality 3D scenes has long been a central challenge in computer graphics, hinging critically on the choice of 3D representation.
Unlike images, 3D scenes must capture not only appearance but also geometry and structural details.
Discrete representations such as meshes and point clouds offer explicit control, enabling direct manipulation and composition~\cite{pointshop}. However, their irregular and heterogeneous nature complicates generation.
To address this, continuous field-based representations like signed distance functions (SDFs)~\cite{deepsdf} and occupancy fields~\cite{occupancy-net, im-net}, with an optional color field for modeling appearance, have been widely adopted for scene generation.
Yet, by enforcing the constraint of a surface, they struggle to model fine details (e.g., fuzzy surfaces, foliage, etc). 
Neural Radiance Fields (NeRFs) have recently revolutionized 3D scene reconstruction by modeling scenes as continuous volumetric representations~\cite{nerf, mipnerf-360}. 
Notably, they can be trained within minutes from casual, real-world video footage~\cite{instant-ngp, plenoxels}.
However, their dense nature poses challenges for editing and recomposition, often requiring auxiliary structures (e.g., cages~\cite{nerfshop, nerf-editing}) or post-processing to merge parts seamlessly~\cite{seamless-nerf}.
3D Gaussian Splatting (3DGS)~\cite{3dgs} alleviates these limitations by representing scenes as collections of fuzzy Gaussians, achieving impressive rendering speed, quality, and robustness to casual inputs. 
Crucially, Gaussians blend naturally, enabling smooth transitions and composability. 
Given these advantages, we adopt 3D Gaussians as our base representation.
To efficiently generate and manipulate them, we employ sparse 3D voxel grids, which abstract away their irregular spatial distribution. 
Sparse voxel grids have long been used as a scalable 3D representation~\cite{gigavoxels} and are increasingly prominent in generative modeling~\cite{xcube, scube, trellis, shapeshifter}.

\paragraph{Controllable 3D scene generation}
Recent advances in 2D generative models~\cite{ldm} and the emergence of large-scale 3D datasets~\cite{objaverse} have significantly accelerated progress in 3D scene generation.
A line of works leverages 2D diffusion priors via Score Distillation Sampling~\cite{dreamfusion, magic3d}, enabling text-driven generation with controllability through bounding boxes~\cite{po-compositional} or automatically learned layouts~\cite{layout-learning}. 
Additional guidance can come from coarse shape inputs~\cite{coin3d}, sketches~\cite{sketchdream, control3d}, or exemplar-based fine-tuning~\cite{themestation}.
While there have been notable gains in efficiency~\cite{latte3d}, fidelity~\cite{sdi}, and the use of multiview priors~\cite{mvdream}, these methods remain computationally intensive and largely limited to simple, object-centric scenes. 
In contrast, recent feedforward approaches conditioned on input images~\cite{trellis, splatter-image} offer faster inference, yet often rely on spherical viewpoint assumptions~\cite{luciddreamer, wonderworld}, produce unrealistic object-level assets~\cite{trellis}, or lack true geometric consistency~\cite{bolt3d}.
Alternative methods condition generation on scene layouts~\cite{blockfusion, scenefactor}, providing greater structural control. However, they typically require domain-specific datasets and extensive supervision, limiting their applicability to constrained environments such as urban landscapes~\cite{citygen, citydreamer} or indoor scenes~\cite{controlroom3d, cc3d}.

\paragraph{Exemplar-based methods}
Exemplar-based methods have a rich history in Computer Graphics, with applications including texture synthesis~\cite{texture-efros,image-quilting}, image analogies~\cite{image-analogies}, and part-based shape modeling~\cite{example-modeling}.
Many of these approaches are grounded in patch-based synthesis or optimization techniques~\cite{geotransfer-patch,survey-patch}. 
Patch-based synthesis algorithms alternate repetitively between two stages: (a) an exact or approximate matching algorithm pairs a patch of the target representation to a patch in the exemplar, (b) patches from the exemplar are aggregated to produce a new target representation. 
A key contribution in this space is the \emph{PatchMatch} algorithm~\cite{patchmatch}, which introduced a fast, randomized method to amortize the patch-matching process.
Sin3DGen~\cite{sin3dgen} adapted this strategy to 3D scene generation, using a hierarchical coarse-to-fine approach to synthesize radiance volumes. 
However, because it relies on patch-based statistics in a purely top-down manner~\cite{singrav}, it struggles to capture high-level semantics and structural coherence. 
Furthermore, it operates densely over the full radiance field, resulting in long generation times (1\textasciitilde 3 minutes).
Recently, 3D Gaussian Splat brushes~\cite{gaussian-brushes} were introduced, enabling interactive brush-based painting from real-world scenes. However, the method remains limited to curve- or stroke-based control, constraining both interaction and generated outputs to a 2.5D appearance.

Another line of work leverages recent deep learning advances by training neural networks on a single exemplar, effectively amortizing inference time through an upfront training phase.
These methods, both in 2D~\cite{sinfusion, singan, sindiffusion} and in 3D~\cite{sin3dm, singleshapegen}, use random augmentations and limited receptive fields to learn generative models capable of producing variations of the input exemplar.
However, with few exceptions~\cite{shapeshifter}, these approaches often require long training times (on the order of hours) and struggle to generate realistic appearance.
This limitation arises from explicitly modeling the output as point clouds~\cite{shapeshifter}, signed distance fields (SDFs)~\cite{sin3dm}, or occupancy grids~\cite{singleshapegen}, which largely confines them to clean, mesh-like geometry and makes it challenging to handle the fuzzy regions often found in real-world scenes.

\section{Method}
We present a method for synthesizing realistic 3D scenes from a single in-the-wild exemplar.
Figure~\ref{fig:overview} illustrates an overview of our pipeline.
Starting from an input video, we reconstruct a 3D Gaussian Splatting scene enriched with semantic features (Section~\ref{ssec:3dgs}).
To obtain a more compact and generation-friendly representation, we abstract the irregular splats into a sparse voxel grid, where each cell encodes a feature vector that combines both appearance and semantic information (Section~\ref{ssec:voxels}).
The grid is synthesized in 0.5-2 seconds using Generative Cellular Automata (GCA), a lightweight generative model conditioned on coarse voxel inputs and trained per scene in less than 10 minutes (Section~\ref{ssec:gca}).
Finally, we recover a photorealistic 3D scene by remapping Gaussians through a sparse, patch-based consistency step (Section~\ref{ssec:patch}).

\subsection{Semantically Augmented 3D Gaussian Splatting}
\label{ssec:3dgs}

3D Gaussian Splatting~\cite{3dgs} represents scenes as a set of anisotropic 3D Gaussians with positions $\boldsymbol{\mu}\in\mathbb{R}^3$ and covariance matrices $\boldsymbol{\Sigma}\in\mathbb{R}^{3\times 3}$ parameterized by a scaling vector $\mathbf{s}\in\mathbb{R}^3$ and a rotation quaternion $\mathbf{r}\in\mathbb{R}^4$.
Their appearance is described by an opacity coefficient $\eta\in\mathbb{R}$ and a view-dependent color \EDIT{$\mathbf{c}:\mathbb{S}^2\to\mathbb{R}^3$} originally parameterized by spherical harmonics with coefficients $\boldsymbol{\gamma}\in\mathbb{R}^m$ (where $m$ is typically $16$ for spherical harmonics of degree $3$).
3D Gaussians are rendered and optimized through a differentiable alpha-compositing rasterization algorithm as follows:
\begin{equation}
    \label{eq:alpha-compositing}
    C = \sum_i \mathbf{c}^{(i)}\alpha^{(i)}\prod_{j=1}^{i-1}(1-\alpha^{(j)}), \;\text{where} \;\alpha^{(i)}=\eta^{(i)}G^{(i)}_\text{2D}(\mathbf{x})
\end{equation}
and $G^{(i)}_\text{2D}(\mathbf{x})$ is the 2D linearized Gaussian density kernel of 3D Gaussian $i$ after projecting it on the 2D viewplane. Please refer to \cite{3dgs} for more details on the rendering algorithm.

To incorporate higher-level scene context and guide generation beyond raw appearance, we draw inspiration from previous works \cite{n3f,feature-3dgs,dff} and augment the unstructured mixture of independent Gaussians with semantic features $\mathbf{f} \in \mathbb{R}^d$.
These features also enable interactive selection of subregions of the scene in our GUI (see Figure~\ref{fig:ui} and the supplemental video).
In practice, we adopt DINO features~\cite{dino, dinov2}. %
Initially extracted as 2D feature maps with 768 channels, we reduce their dimensionality to 8 via PCA, and lift them to 3D by replacing $\mathbf{c}^{(i)}$ with $\mathbf{f}^{(i)}$ in Equation~\eqref{eq:alpha-compositing}.
Additional implementation details are provided \EDIT{in the supplemental material (Section S.2.1).} %

\subsection{Sparse Voxel Grids}
\label{ssec:voxels}

In order to facilitate and accelerate scene generation, we abstract the set of 3D Gaussians in a sparse voxel grid $\voxelset=\{\mathbf{p}_i\}_{i=1}^N$ where $N$ is the number of voxels and $\mathbf{p}_i\in\mathbb{Z}^3$ is the positional index of an occupied voxel in 3D.
Each voxel is enriched with a feature vector $\mathbf{z}_i$, forming a sparse feature volume $\featureset=\{\mathbf{z}_i\}_{i=1}^N$ that captures both the semantic and appearance attributes of the 3D Gaussians it encompasses.
This process is illustrated in Figure~\ref{fig:overview} (a).

More precisely, we instantiate voxels where 3D Gaussians exceed a predefined opacity threshold $\tau=\num{0.1}$.
We build the volume of voxels at a fine resolution $r_t=64^3$, and we derive a coarse volume at resolution $r_c=16^3$ by downsampling the fine volume.
The coarse volume is used as a conditioning signal (see Section~\ref{ssec:gca}), while we synthesize the scene at the more expressive fine resolution.
\EDIT{The supplemental material (Section S.3.1)} discusses the trade-offs between resolution choices.
We maintain diversity among cells by selecting the features of the Gaussian with the highest opacity rather than simply averaging features from all Gaussians.
We extract both the appearance encoded in the coefficients of spherical harmonics of the chosen splat and the semantic information of the distilled DINO features presented in Section~\ref{ssec:3dgs}. 
In practice, we use four dimensions for each of them, resulting in an 8-dimensional feature per voxel. 
For consistency, each of them is independently PCA-ed and subsequently re-normalized.
Additional details are provided \EDIT{in the supplemental material (Section S.2)}.

\subsection{Generating Sparse Voxels with GCA}
\label{ssec:gca}

To generate sparse voxels grids, we adapt \emph{Generative Cellular Automata} (GCA)~\cite{gca, cgca}.
At a high level, GCA only predicts occupied surface voxels and their corresponding features as described in the previous section.

\subsubsection{Background: \textit{Generative Cellular Automata}}
\label{ssec:gca-background}

\begin{figure}
    \centering
    \includegraphics[width=0.9\linewidth]{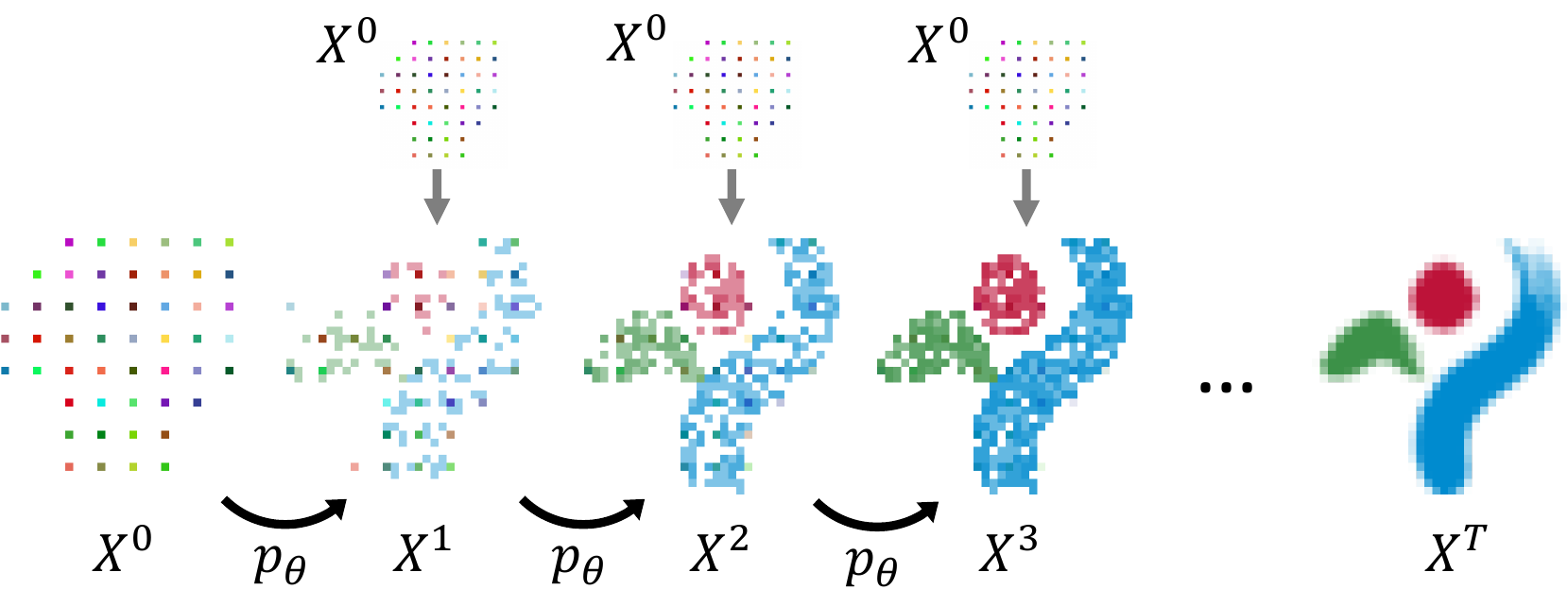}  
    \caption{
        At each time step $t$, GCA samples a new state $\gcastate^{t + 1}$ composed of sparse voxel occupancies equipped with features from $p_{\theta}(\gcastate^{t+1}\probcond\gcastate^t, \gcastate^0)$. 
        Recursively applying the transition kernel $p_{\theta}$ to the initial state $\gcastate^0$ yields the final generated state $\gcastate^T$.
        We adapt GCA to conditional generation by choosing $\gcastate^0$ to be the upsampled coarse set of conditioning voxels and concatenating it to $\gcastate^t$ each time $p_\theta$ is applied (gray arrows on the top).
    }
    \label{fig:gca}
\end{figure}

GCA~\cite{gca, cgca} represents shapes as a sparse grid of voxels $\voxelset$ equipped with per-voxel features $\featureset$, effectively modeling the joint distribution $\gcastate = (\voxelset, \featureset)$.
As illustrated in Figure~\ref{fig:gca}, starting from an initial state $\gcastate^0 = (\voxelset^0, \featureset^0)$, GCA generates a complete shape by iteratively sampling intermediate states $\gcastate^t= (\voxelset^t, \featureset^t)$ until reaching a final state $\gcastate^T$:
\begin{equation*}
    \gcastate^{t + 1} \sim p_\theta(\cdot\probcond \gcastate^t), \label{eq:transition_kernel}
\end{equation*}
where $T$ is the number of steps and $p_\theta$ is a learnable transition kernel.
This kernel is implemented as a U-Net~\cite{unet} with sparse convolutions, and only updates cells in the neighborhood of already occupied cells via local update rules similar to cellular automata.
This inductive bias exploits the inherent connectivity and sparsity of 3D shapes, drastically reducing the search space in a high‐resolution grid.
For tractability, the kernel factorizes the joint probability of a state into separate terms for occupancy and features:
\begin{equation}
    p_\theta(\gcastate^{t + 1} \probcond \gcastate^{t}) = \underbrace{p_\theta(\voxelset^{t+1}\probcond \gcastate^t)}_{(a)}\, \underbrace{p_\theta(\featureset^{t+1}\probcond\gcastate^t, \voxelset^{t+1})}_{(b)}
\end{equation}
where $(a)$ is modeled as a Bernoulli distribution and $(b)$ as a Gaussian distribution whose uncertainty decreases over time.
GCA is trained through a process called \emph{Infusion}~\cite{infusion}.
\EDIT{The supplemental material (Section S.1)} provides for a full description of GCA and a discussion of its training strategy.

\subsubsection{Single-exemplar GCA and Controllable Generation}
\label{ssec:controllable-gca}

We adapt the generative framework of GCA to learn diverse yet plausible variations from a single input exemplar. 
To do so, we use a shallow network, greatly reducing the receptive field, drawing inspiration from prior work on single-image generative models~\cite{sinfusion, sindiffusion}.
\EDIT{
In addition, following prior work~\cite{sin3dm}, we introduce random augmentations to enable the model to capture local structure more flexibly.
Concretely, we use random voxel crops at fine resolution, each with side lengths between 25 and 30 and containing at least 250 voxels in total (via rejection sampling).
To prevent GCA from learning the boundaries of these crops, we pad each crop by 2 voxels along every dimension; these padded voxels are not used to supervise the transition kernel but are provided as contextual input to the network.
}
Implementation details are provided in \EDIT{the supplemental material (Section S.2.2)} and ablation studies are discussed in Section~\ref{ssec:ablations}.

Another key objective is to enable control over scene generation. 
Starting from the volume of voxels described in Section~\ref{ssec:voxels}, we use its downsampled coarse version $r_c$ both as the initial state $\gcastate^0$ and as a conditioning signal at every step (see Figure~\ref{fig:gca}). 
During training, the conditioning volume $r_c$ is derived from the ground-truth scene (i.e., via teacher forcing), while at inference time, it can be replaced with any valid voxel input. 
Note that generation occurs at the higher and more expressive resolution $r_t$, starting from features randomly initialized with a standard normal distribution.
As discussed and ablated in \EDIT{the supplemental material (Section S.2.1)}, the choice of resolutions for $r_c$ and $r_t$ is critical for balancing controllability and output diversity.

\subsection{Patch-based Consistent Scene Composition} 
\label{ssec:patch}

\begin{figure}
    \centering
    \includegraphics[width=0.95\linewidth]{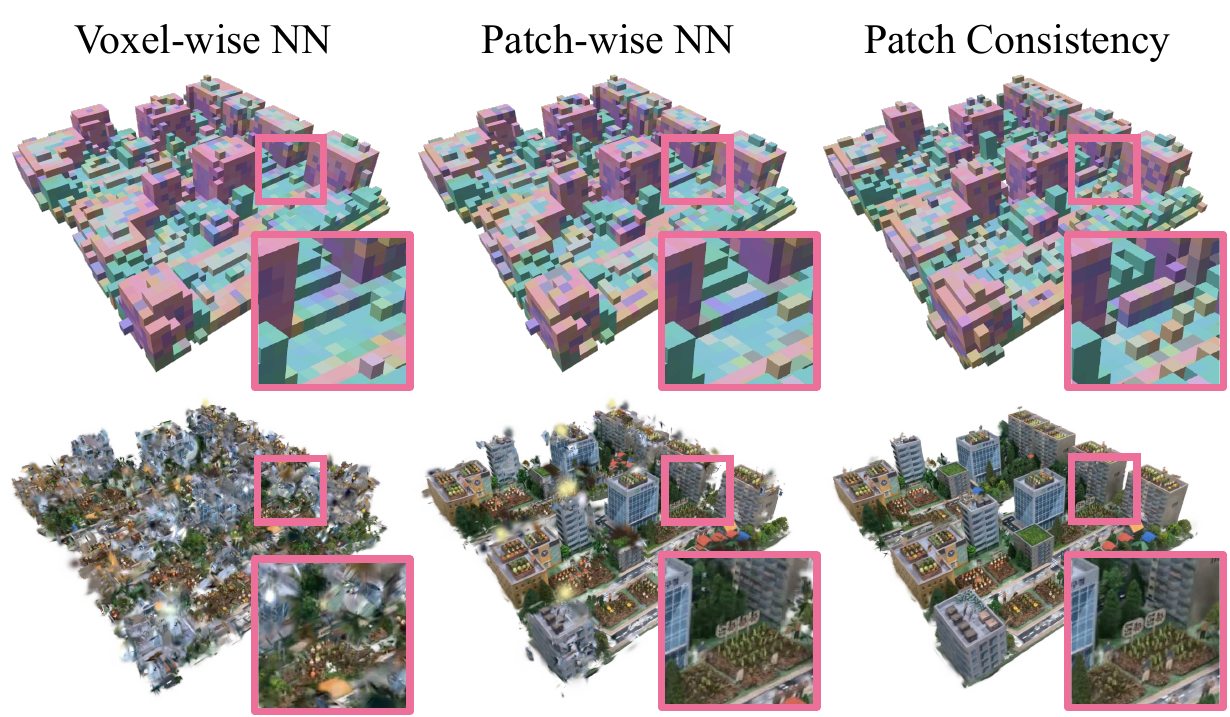}  
    \caption{
        \textit{Voxel-wise NN}.
        Na\"ively remapping each predicted voxel to its nearest feature in the exemplar and filling in the corresponding 3D Gaussians fails to produce consistent results due to local consistency and the approximate predictions of GCA. 
        \textit{Patch-wise NN}.
        Using the patch-wise distance defined in Equation~\eqref{eq:patch-dist} improves visual coherence but still fails to account for modeling errors. 
        \textit{Patch Consistency}.
        We thus propose an additional sparse patch-based consistency operation to refine missing local statistics.
    }
    \label{fig:patchmatch}
\end{figure}

A final patch-based consistency step converts the generated low-dimensional voxels into a scene with a rich photorealistic appearance composed of 3D Gaussian splats extracted from the input exemplar.
If we na\"ively copy and paste Gaussians in the cell with the closest feature vector from the exemplar scene, individual voxels are agnostic of their neighborhood, resulting in poor spatial consistency.
We denote this approach as ``Voxel-wise NN'', and Figure~\ref{fig:patchmatch} contains an example.
We can enforce more consistency by considering a larger patch-wise context when selecting which cell to borrow the Gaussians from, denoted as ``Patch-wise NN''. However, this method breaks down when the voxels synthesized by GCA deviate too much from the input exemplar.
To address this issue, we draw inspiration from patch-based image synthesis methods~\cite{survey-patch}, introducing an efficient \emph{sparse} patch-based consistency operation that redistributes the patch-wise statistics of the exemplar into a coherent spatial arrangement.

Patch-based synthesis was originally developed for dense 2D image domains, where simple per-pixel metrics can be directly applied. In contrast, extending this idea to 3D requires accounting for voxel occupancy to track sparse geometry. 
We adapt it by defining a distance directly over sparse voxels, combined with a \emph{voting mechanism} to locally grow or ungrow geometry moderately.

\paragraph{Notations and patch extraction.}
\EDIT{
We refine the GCA output
$\gcastate^0=(\voxelset^0,\featureset^0)$
into a final state
$\gcastate^K=(\voxelset^K,\featureset^K)$
through an iterative patch-based synthesis procedure, before reconstructing
a photorealistic 3D Gaussian scene.
Let $\voxelset^k\subset\mathbb{Z}^3$ denote the occupied voxel coordinates at iteration $k$ of this process, and
$\featureset^k=\{\mathbf{z}_{\mathbf{p}}^k\}_{\mathbf{p}\in\voxelset^k}$
the corresponding voxel features.
We define dense indicator and feature volumes:
\begin{equation*}
O^k(\mathbf{p})=\mathbb{I}[\mathbf{p}\in\voxelset^k],\qquad
F^k(\mathbf{p})=
\begin{cases}
\mathbf{z}_{\mathbf{p}}^k,&\mathbf{p}\in\voxelset^k,\\
\mathbf{0},&\text{otherwise}.
\end{cases}
\end{equation*}
Note that we use $k$ instead of $t$ to distinguish these iterative steps from the sampling steps of GCA.
}

\EDIT{
Let $\mathcal{U}=\{-\tfrac{l-1}{2},\dots,\tfrac{l-1}{2}\}^3$ be the set of
patch offsets with $p=l^3$.
For any voxel center $\mathbf{p}$ and state $\gcastate^k$, we define the
associated patch
\begin{equation*}
\begin{aligned}
P_{\gcastate^k}(\mathbf{p})
&=
\bigl(O^k_{\mathbf{p}}, F^k_{\mathbf{p}}\bigr),\\
O^k_{\mathbf{p}}[\mathbf{u}]
&= O^k(\mathbf{p}+\mathbf{u}), \qquad
F^k_{\mathbf{p}}[\mathbf{u}]
= F^k(\mathbf{p}+\mathbf{u}) .
\end{aligned}
\end{equation*}
for all $\mathbf{u}\in\mathcal{U}$.
We denote by $\gcastate^E$ the exemplar voxels.
}

\paragraph{Patch distance and matching.}
\EDIT{
Given two patches $P_e$ (resp.\ $P_g$) in the exemplar (resp.\ in the generated scene), we measure their similarity by
\begin{equation}
d(P_e,P_g)=(1-w)\,d_{\mathrm{occ}}(P_e,P_g)
+w\,d_{\mathrm{feat}}(P_e,P_g),
\label{eq:patch-dist}
\end{equation}
where
\begin{align*}
d_{\mathrm{occ}}
&=
1-\frac{1}{p}
\sum_{\mathbf{u}\in\mathcal{U}}
O_e[\mathbf{u}]\,O_g[\mathbf{u}],\\[3pt]
d_{\mathrm{feat}}
&=
1-
\frac{
\sum_{\mathbf{u}\in\mathcal{U}}
O_e[\mathbf{u}]\,O_g[\mathbf{u}]
\langle F_e[\mathbf{u}],F_g[\mathbf{u}]\rangle
}{
\sum_{\mathbf{u}\in\mathcal{U}}
O_e[\mathbf{u}]\,O_g[\mathbf{u}]
}.
\end{align*}
}
\EDIT{
At iteration $k$, each occupied voxel center
$\mathbf{p}\in\voxelset^k$ selects its best-matching exemplar patch center
\begin{equation}
\phi^k(\mathbf{p})
=
\arg\min_{\mathbf{q}\in\voxelset^E}
d\!\left(
P_{\gcastate^E}(\mathbf{q}),
P_{\gcastate^k}(\mathbf{p})
\right).
\label{eq:patch-assign}
\end{equation}
}

\paragraph{Patch aggregation and voting.}
\EDIT{
Each assignment $\phi^k(\mathbf{p})$ proposes a new exemplar patch.
For any voxel $\mathbf{r}$ covered by the patch centered at $\mathbf{p}$,
we define
\begin{align*}
\tilde O_{\mathbf{p}}^k(\mathbf{r})
&=
O^E\!\big(\phi^k(\mathbf{p})+(\mathbf{r}-\mathbf{p})\big),\\
\tilde F_{\mathbf{p}}^k(\mathbf{r})
&=
F^E\!\big(\phi^k(\mathbf{p})+(\mathbf{r}-\mathbf{p})\big).
\end{align*}
}

\EDIT{
Let
$\mathcal{P}(\mathbf{r})=
\{\mathbf{p}\in\voxelset^k : \mathbf{r}\in\mathbf{p}+\mathcal{U}\}$
be the set of patch centers whose patches overlap $\mathbf{r}$.
We update voxel features by occupancy-weighted averaging,
\begin{equation*}
F^{k+1}(\mathbf{r})
=
\frac{
\sum_{\mathbf{p}\in\mathcal{P}(\mathbf{r})}
\tilde O_{\mathbf{p}}^k(\mathbf{r})\,
\tilde F_{\mathbf{p}}^k(\mathbf{r})
}{
\sum_{\mathbf{p}\in\mathcal{P}(\mathbf{r})}
\tilde O_{\mathbf{p}}^k(\mathbf{r})
},
\label{eq:patch-blend}
\end{equation*}
and compute an occupancy vote
\begin{equation*}
v^k(\mathbf{r})
=
\frac{1}{|\mathcal{P}(\mathbf{r})|}
\sum_{\mathbf{p}\in\mathcal{P}(\mathbf{r})}
\tilde O_{\mathbf{p}}^k(\mathbf{r}).
\end{equation*}
We keep voxels whose occupancy vote exceeds $\beta$ (set to 0.5 in practice).
We repeat this procedure (i.e., matching, followed by aggregation and voting) for $K$ iterations.
}

\paragraph{Reconstruction of a 3D Gaussian scene.}
\EDIT{
After $K$ iterations, each synthesized occupied voxel $\mathbf{p}\in\voxelset^K$ selects a single voxel from the exemplar using the distance in Equation~\eqref{eq:patch-dist} via the mapping in Equation~\eqref{eq:patch-assign}.
All Gaussians belonging to the selected exemplar voxel are then copied to the voxel corresponding to $\mathbf{p}$, without any further processing.
}

\paragraph{Discussion}
As shown in Figure~\ref{fig:patchmatch}, this procedure is key to recover missing geometric details.
To allow only small refinements from the initial generated voxels, we limit additional voxels from being added farther than a distance $\lambda_\text{patch}$. 
We use a patch size of $l=\num{5}$, $\num{7}$ iterations and $\lambda_\text{patch}=2$. 
\EDIT{
Note that the distance in Equation~\eqref{eq:patch-dist} may assign high values to patches that are similar yet have low occupancy. 
In practice, we observed only minor adverse effects, as patches are updated for a limited number of iterations and the extent of geometric changes is restricted through $\lambda_{\text{patch}}$.
}
\EDIT{The supplemental material (Section S.3.2)} presents a discussion and ablation studies of these parameter choices.
\EDIT{Section S.4 of the supplemental material} highlights the differences between our approach and the related method Sin3DGen~\cite{sin3dgen}.

\begin{figure}
    \centering
    \includegraphics[width=\linewidth]{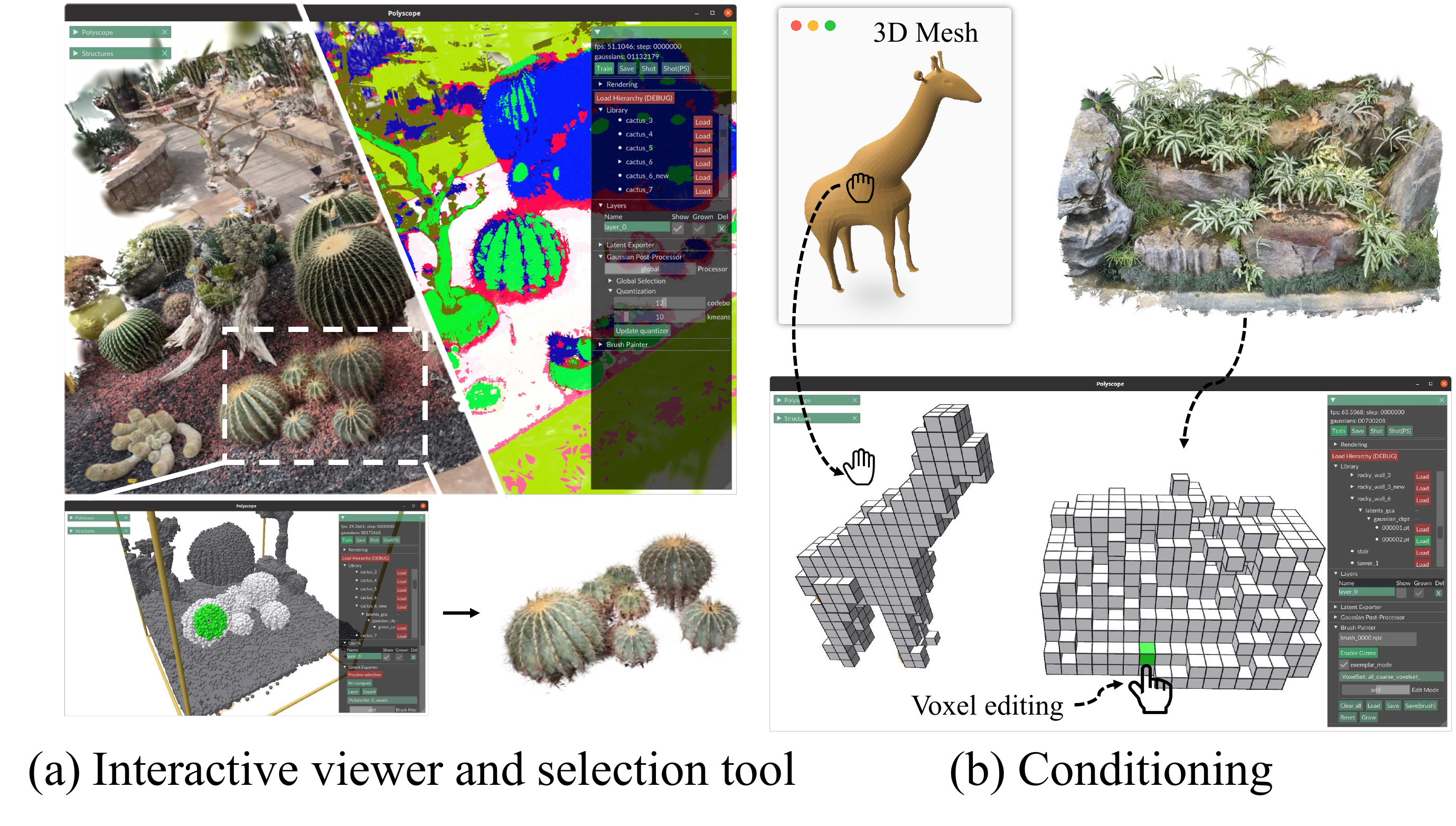}
    \caption{
    Illustration of our user interface. 
    (a) Our interactive viewer runs at real-time framerates (30-60fps) and comes with a selection tool using the quantized DINO features. Additional adjustments can be made at any stage with a manual selection tool. 
    (b) Conditioning can be performed using parts of the exemplar, voxelized 3D meshes, or by manually editing voxels.
    }
    \label{fig:ui}
\end{figure}

\section{Results}

\begin{figure*}
    \centering
    \includegraphics[trim={0, 3mm, 0, 5mm}, clip, width=\textwidth]{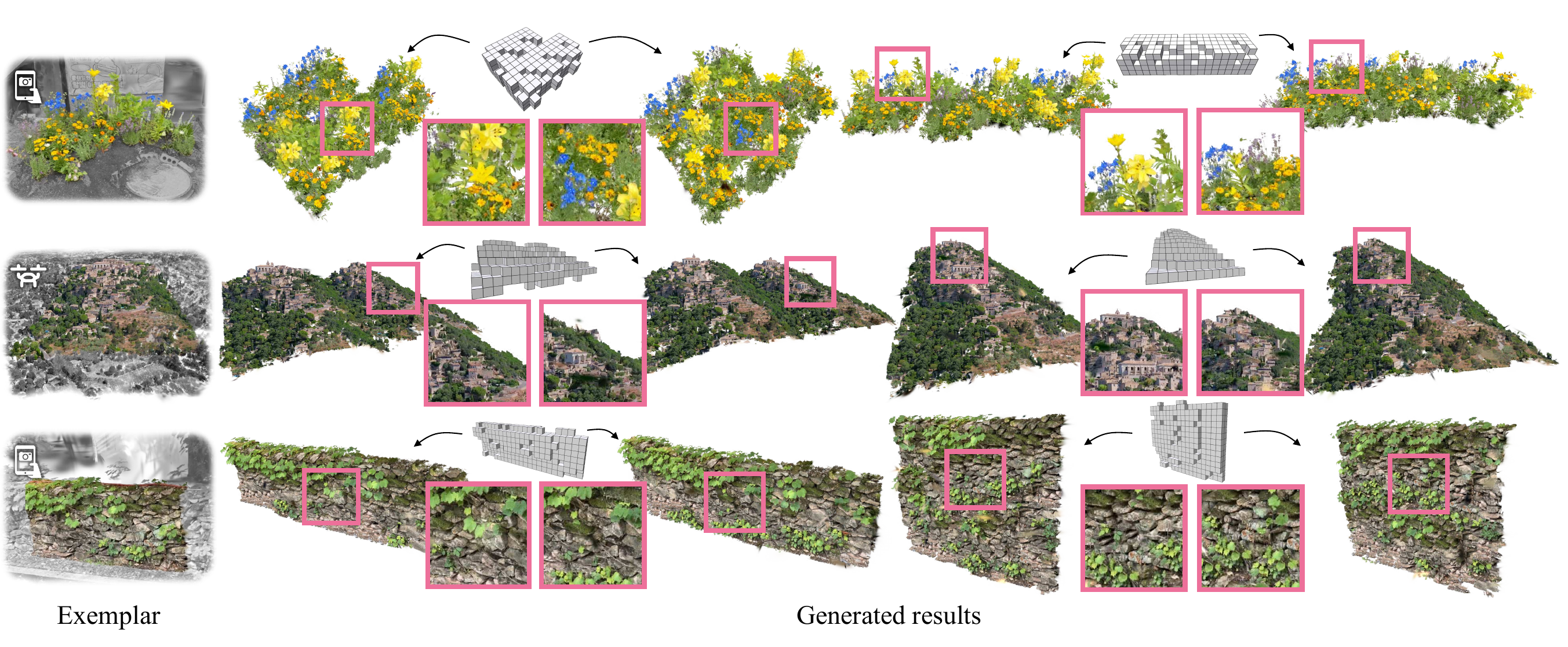}
    \caption{ 
        Starting from the exemplar on the left, we generate multiple samples for different conditioning signals.
        All examples are derived from real-world scenes obtained from casual mobile footage (rows 1, 3) or a drone (row 2).
    }
    \label{fig:diverse_results_in_text}
\end{figure*}

We demonstrate our method’s ability to generate high-quality and controllable 3D scenes from casually captured in-the-wild videos.
First, we introduce our interactive user interface (Section~\ref{ssec:user-interface}), which allows users to import videos, select regions of interest, and author novel 3D scenes through intuitive composition of multiple exemplars.
We then demonstrate the visual quality and controllability of generated scenes in diverse real-world environments (Section~\ref{ssec:generations}).
Next, we ablate key components of our method (Section~\ref{ssec:ablations}).
Finally, we analyze the speed, showcasing its responsiveness and efficiency for interactive use (Section~\ref{ssec:speed}).
Full implementation details are available in \EDIT{the supplemental material (Section S.2)}. %

\subsection{Interactive User Interface}
\label{ssec:user-interface}

We provide an interactive user interface that enables users to import videos, select regions of interest, and generate novel 3D scenes, as shown in Figure~\ref{fig:ui}.
Real-world scenes often contain cluttered backgrounds and lack clear object boundaries, motivating the need for an intuitive selection mechanism.
To address this, we apply k-means clustering to the semantic features of 3D Gaussians, allowing users to interactively adjust the number of clusters and refine the selection through annotation or direct manipulation of 3D Gaussians (Figure~\ref{fig:ui}~(a), \EDIT{Figure S.2 in supplemental material} and supplemental video).

We use a sparse voxel grid to provide the conditioning signal for generation, as illustrated in Figure~\ref{fig:ui} (b).
In practice, the coarse geometry can be obtained from 1) coarse voxels from the existing exemplar, 2) input meshes, and/or 3) direct voxel editing.
After generating each asset individually using these coarse voxels, our UI facilitates compositing them into a complete scene (see the supplemental video).

\subsection{Scene Generation}
\label{ssec:generations}

\begin{figure*}
    \includegraphics[width=1.0\textwidth]{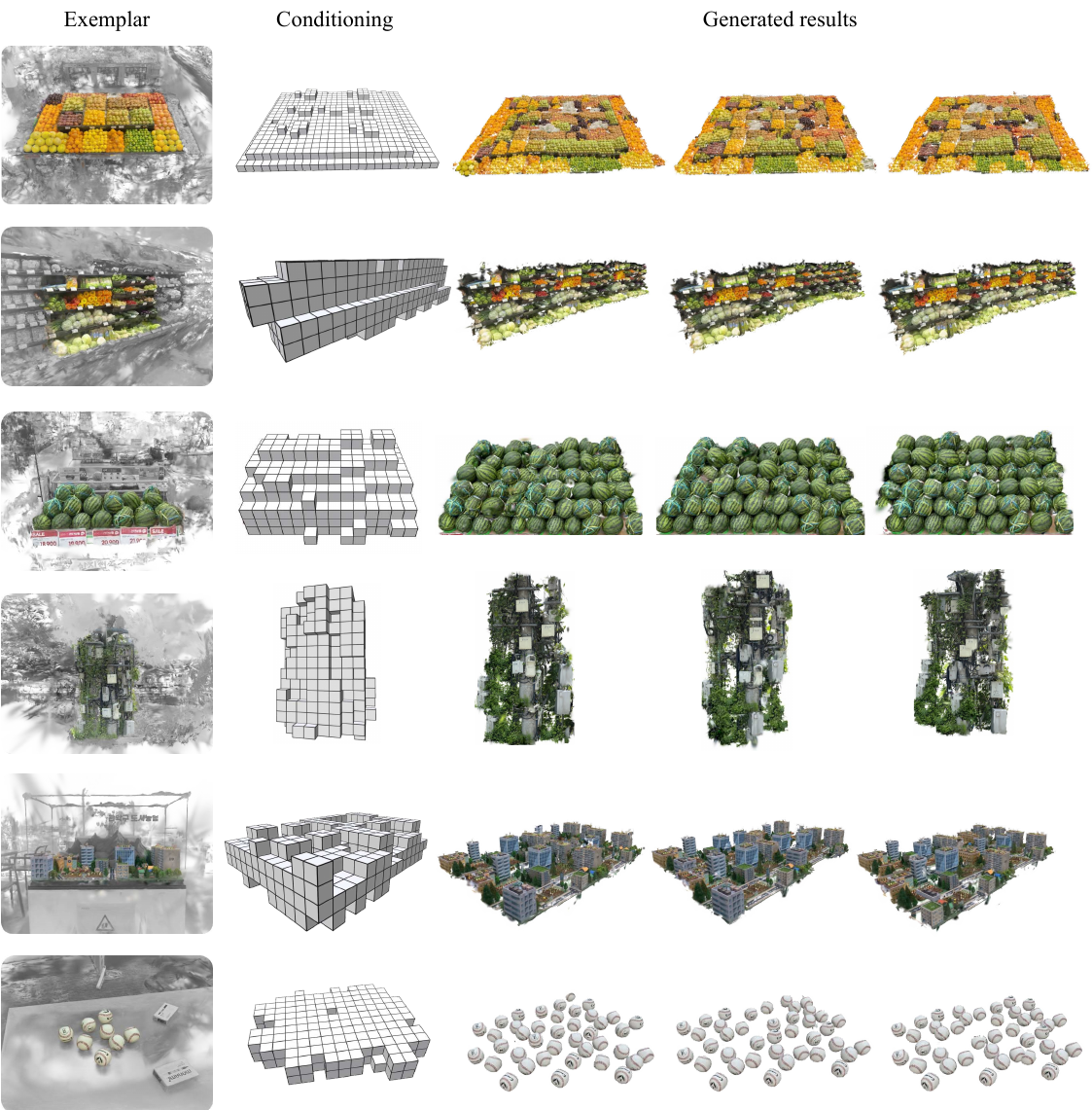}
    \caption{
    From a single real-world exemplar (column 1), our method can generate diverse results (columns 3-5) conditioned on coarse voxel inputs (column 2). 
    Generation is performed and visualized within 0.5-2 seconds in our interactive editor, enabling users to iteratively refine the conditioning input to produce the target asset.
    }
    \label{fig:diverse_results_fig_only}
\end{figure*}

We demonstrate that our method can generate diverse, controllable, and high-quality 3D scenes from casually captured real-world videos.
All input videos are sourced from the LeRF dataset~\cite{lerf}, YouTube, or recorded by ourselves, ensuring they reflect unconstrained real-world environments rather than synthetic or curated scenes.
Figures~\ref{fig:diverse_results_in_text} and~\ref{fig:diverse_results_fig_only} showcase examples generated from a single exemplar video conditioned on coarse voxel inputs.
Given the same conditioning input, our method produces multiple plausible and visually distinct scenes within seconds, demonstrating its capacity for interactive and diverse content generation.  %

\begin{figure}
    \centering
    \includegraphics[trim={0, 3mm, 0, 0}, clip, width=0.8\linewidth]{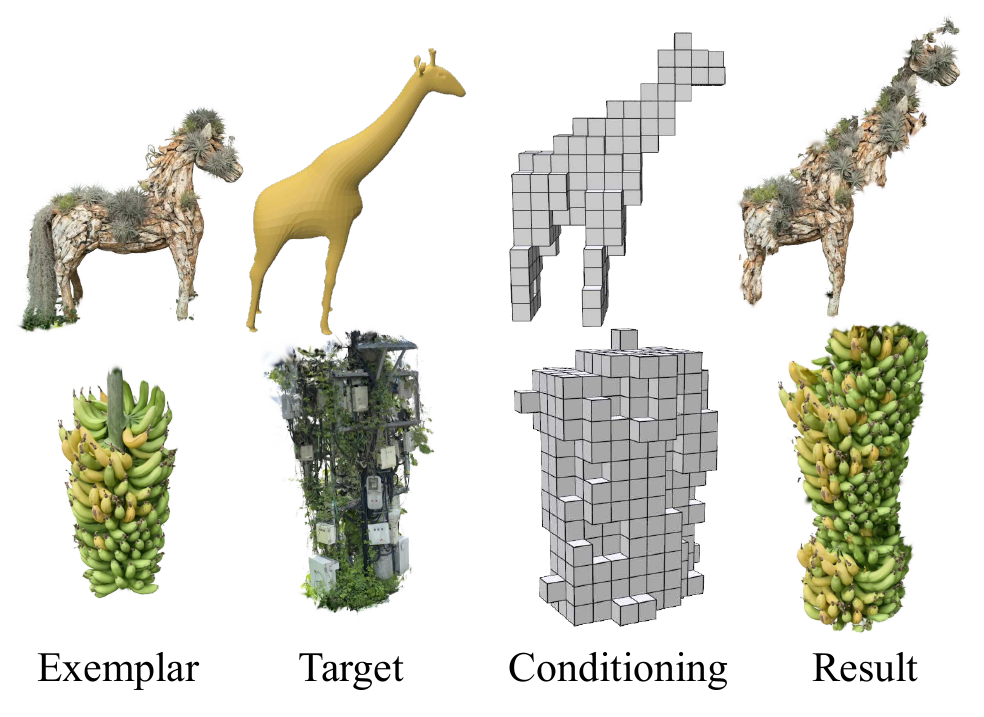}
    \caption{
        Our method can be used to transfer the appearance of an exemplar to a mesh (row 1) or another exemplar (row 2). Given an input exemplar (column 1), and a target geometry (column 2), the target geometry is converted to conditioning voxels (column 3), and our method generates a new result (column 4).}
    \label{fig:analogy}
    \vspace{-0.5em}
\end{figure}

Beyond diversity, our method offers explicit controllability.
Users can guide the synthesis process using coarse voxel geometry derived from exemplars, meshes, or manual edits, and can further manipulate results by transferring appearance to novel geometries, either from a mesh or another exemplar scene (Figure~\ref{fig:analogy}).
This flexibility enables users to reconfigure and remix scene elements with minimal effort (see the supplemental video).

\subsection{Ablations}
\label{ssec:ablations}

\begin{figure}
    \centering
    \includegraphics[width=\linewidth]{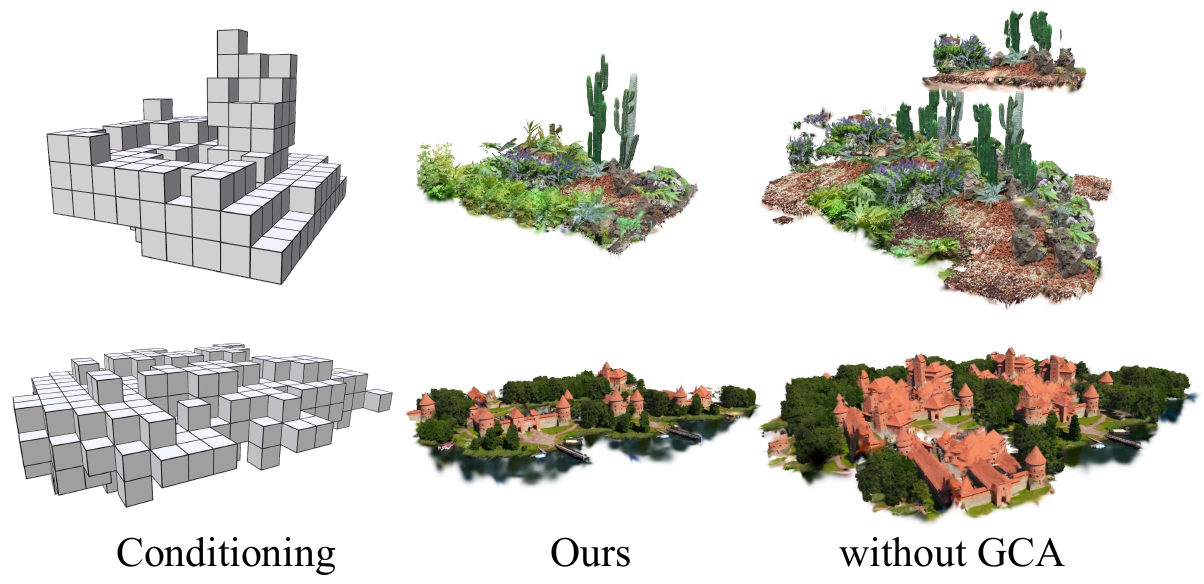}
    \caption{
    Removing GCA during generation and relying solely on our sparse patch-based synthesis (column 3) leads to repetitive patterns, diminished spatial and semantic consistency, and geometry that exceeds the conditioning voxels (column 1), in contrast to our two-stage strategy with GCA (column 2).
    }
    \label{fig:full_patchmatch}
\end{figure}

A distinctive design choice of our method is that it operates entirely sparsely: we first grow a sparse set of voxels using GCA (Section~\ref{ssec:gca}), followed by local refinements via a sparse patch-based consistency step (Section~\ref{ssec:patch}).
Figure~\ref{fig:full_patchmatch} shows the impact of removing GCA from this pipeline.
More precisely, we generate voxels hierarchically using our sparse patch-based optimization strategy. 
The coarsest resolution ($r_c = 16^3$) is initialized with the conditioning voxels (column 1) and random features. At each level, we apply the same parameters as described in Section~\ref{ssec:patch}, and upsample the volume by a factor of 2.
This approach leads to repetitive patterns, lacks spatial and semantic consistency, and tends to inflate geometry beyond the conditioning input.

\begin{figure*}
    \centering
    \includegraphics[width=0.95\textwidth]{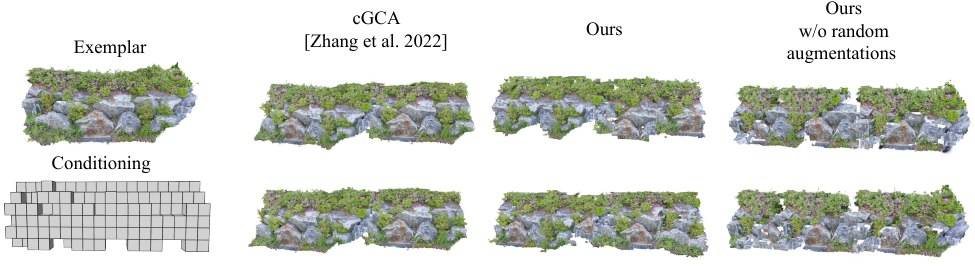}
    \caption{
        We propose a lightweight alternative to cGCA~\cite{cgca}, with a reduced receptive field (see \EDIT{supplemental material, Section S.2.2}).
        Given the exemplar and conditioning input (left), we generate two samples per model.
        cGCA tends to overfit the input exemplar, even with random augmentations.
        Without random augmentations, our model exhibit low sample diversity and introduce discontinuities (right-most column).
    }
    \label{fig:ablation_architecture}
\end{figure*}

A crucial design choice to train GCA from a single exemplar comes from the architecture of the network and the use of random augmentations to enhance the diversity of generated results despite training with a limited
set of inputs.
As shown in Figure~\ref{fig:ablation_architecture}, naively using the deep U-Net architecture proposed by Zhang \etal~\cite{cgca}, denoted as cGCA, results in plain overfitting on the input shape and appearance. 
\begingroup
\sisetup{round-mode = places, round-precision = 2, round-integer-to-decimal}
We thus introduce a smaller network with a more restricted receptive field (details in \EDIT{the supplemental material, Section S.2.2}), which also significantly reduces model size (\SI{4.209692955}{\mega\byte} vs.\ \SI{468.1949692}{\mega\byte}), training (\SI{6.254148678}{\min} vs.\ \SI{23.39996537}{\min} for Figure~\ref{fig:ablation_architecture}) and inference speed (\SI{199.9761105}{\ms} vs. \SI{328.7445068}{\ms} for Figure~\ref{fig:ablation_architecture}). 
\endgroup
Additional ablations for our patch-based consistency step are provided in \EDIT{the supplemental material (Section S.3)}.

\subsection{Training and Generation Speed}
\label{ssec:speed}
\begin{figure}
    \includegraphics[width=\linewidth]{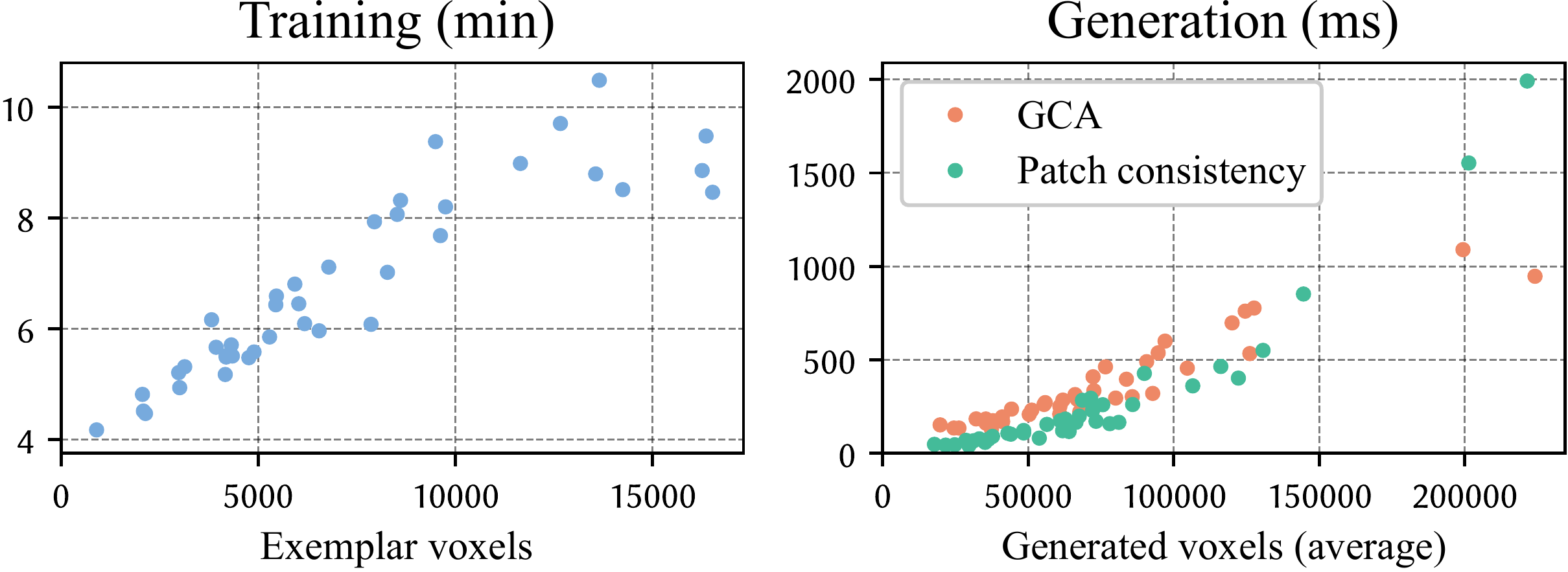}
    \caption{
    Timings for training (left) and generation (right).
    We report training times for 40 manually captured scenes covering various distributions (left). Even for a large number of voxels in the input exemplar, training time is almost always below 10 minutes. 
    Conditioned on each manual input, we generate 10 samples and report the average generation time for GCA (red, right) and the subsequent patch consistency operation (green, right).
    Generation in sum takes less than 2 seconds in most cases.
    }
    \label{fig:ablation_performance_training}
    \vspace{-1em}
\end{figure}

A key advantage of our method is its ability to generate high-quality 3D scenes with minimal latency, enabling interactive use.
Our method incurs low and fixed training overhead.
As shown in Figure~\ref{fig:ablation_performance_training} (left), training the generative kernel from a single exemplar takes under 10 minutes, even for complex inputs.
Once trained, our method supports real-time generation, enabling iterative scene design (see the supplemental video).
As shown in Figure~\ref{fig:ablation_performance_training} (right), our two-stage pipeline (i.e., GCA generation followed by our patch-based consistency step) generates scenes in under 2 seconds, even for large scenes.

\section{Limitations}

\begin{figure}
    \includegraphics[width=\linewidth]{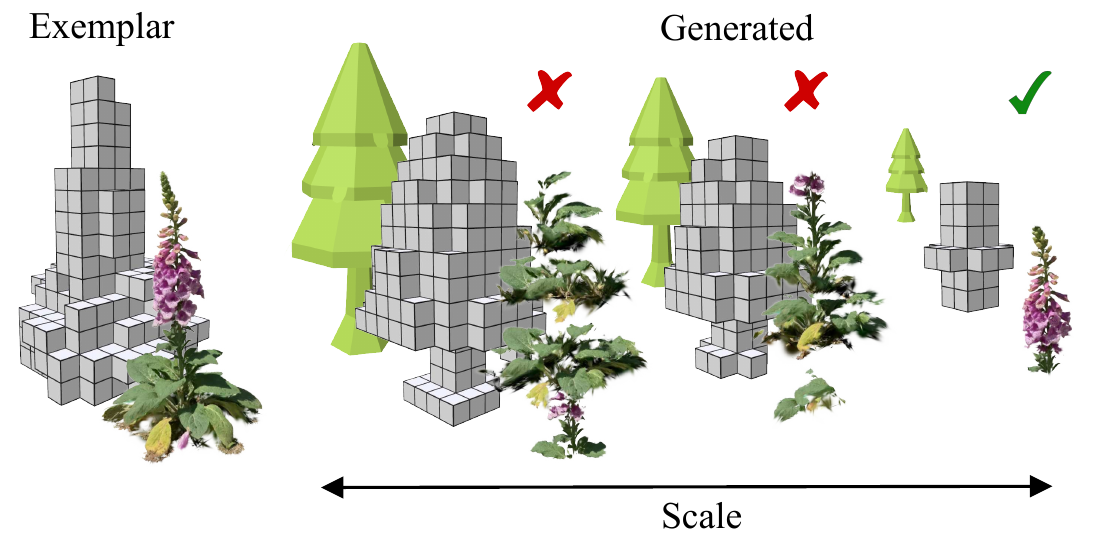}
    
    \caption{From a given input exemplar on the left, we generate a shape given the same input mesh at different scales. Note that the mesh is always voxelized to condition generation. 
    If the input geometry deviates significantly from the structural distribution of the exemplar, our method struggles to produce consistent results.
    }
    \label{fig:ablation_distribution}
\end{figure}

\begin{figure}
    \includegraphics[width=\linewidth]{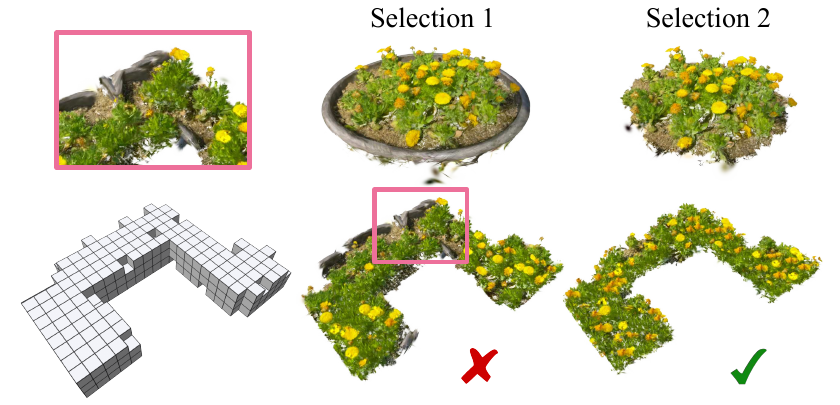}
    
    \caption{
    With two different selections of the same scene, generation can be severely hindered by structural artifacts that lie beyond the invariances that can be modeled by GCA as it operates on axis-aligned voxel grids.
    }
    \label{fig:ablation_selection}
\end{figure}

\begin{figure}
    \includegraphics[width=\linewidth]{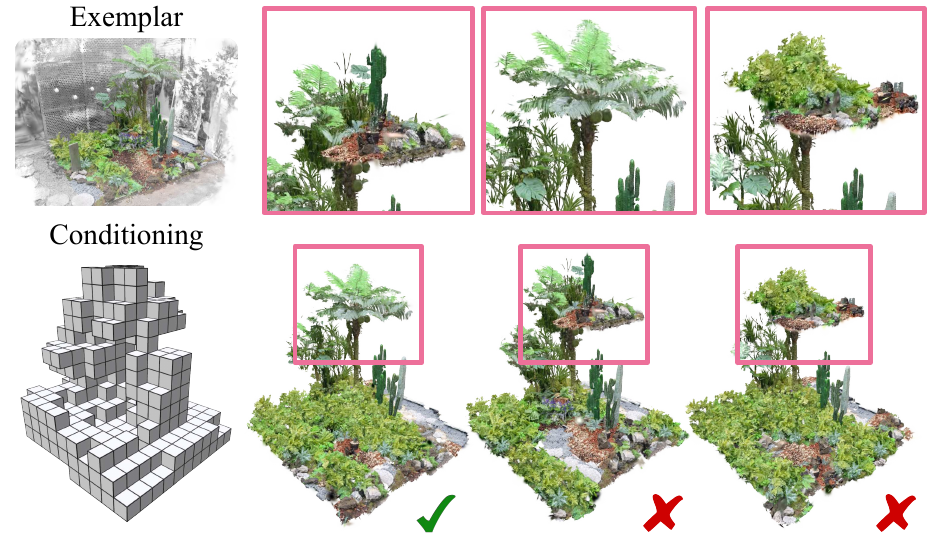}
    \caption{
    Due to its limited receptive field and capacity, our model can sometimes confuse distinct regions during generation, especially when semantically similar areas share the same coarse geometry.}    
    \label{fig:limitations_context}
\end{figure}

\begin{figure}
    \includegraphics[width=\linewidth]{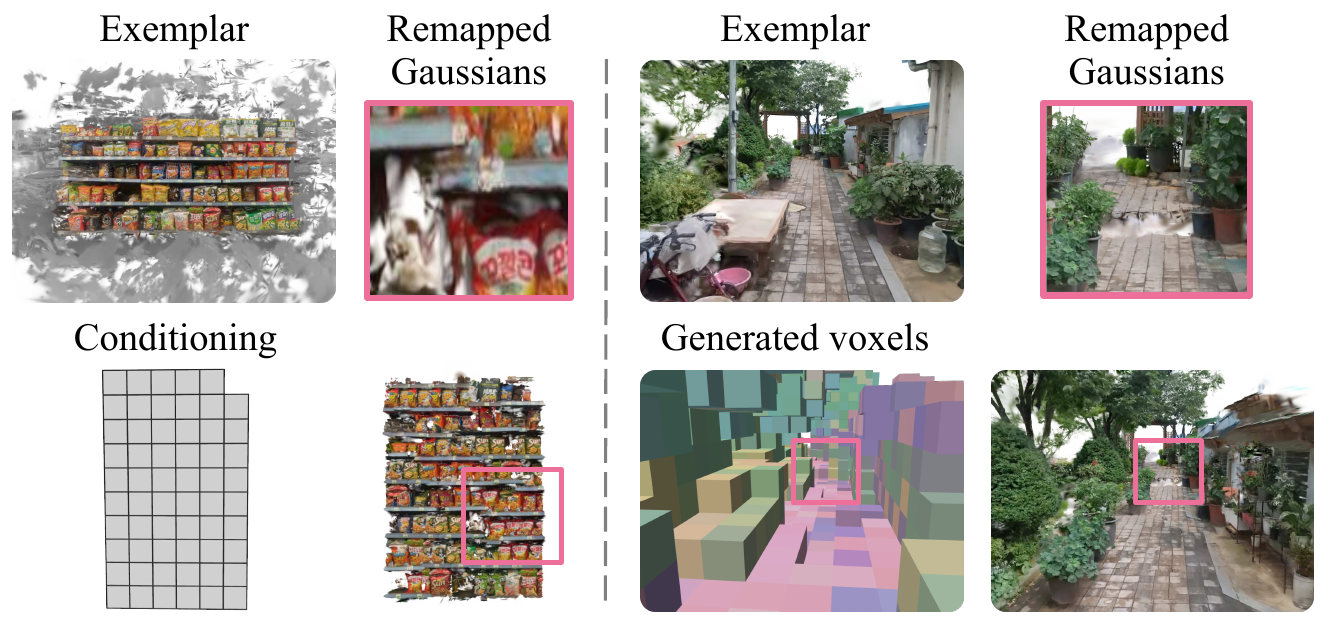}
    \caption{Our method operates on fixed-size voxels that can only be retrieved from the set of voxels in the exemplar. For axis-aligned structured scenes (left) and large scenes (right), this comes with noticeable artifacts, such as misalignments (left) and ``cracks'' (right).}
    \label{fig:limitations_alignment}
\end{figure}

Being trained on a single exemplar without external priors, our generative model struggles to extrapolate to inputs that deviate significantly from its original distribution. This limitation is illustrated in Figure~\ref{fig:ablation_distribution}.
For the same reason, it is also crucial to carefully select the target distribution of the scene. 
As shown in Figure~\ref{fig:ablation_selection}, properly isolating flowers from their pot allows to brush a new scene without artifacts. 
As generation is only conditioned through coarse voxels, ambiguities appear for complicated scenes where the same coarse geometry may be partially repeated throughout the scene. Since our GCA cannot observe the full generation, this results in ``out of place'' generated results as illustrated in Figure~\ref{fig:limitations_context}.
\EDIT{Furthermore, after reconstructing the 3D Gaussians from voxels, no further optimization is performed, which may compromise the quality of the synthesized results, especially causing artifacts at patch boundaries as shown in Figure~\ref{fig:limitations_alignment}.}

The resolution of the conditioning signal is prone to the trade-off between the diversity of synthesized scenes and the amount of control that can be provided to users.
We provide additional analysis in \EDIT{the supplemental material (Section S.3.1).} %
Future work could explore automatic resolution and scale selection or, inspired by recent multi-scale methods~\cite{patchcomplete, xcube}, jointly train nested models and select the one matching the user’s desired level of detail.

We build on the original form of 3D Gaussian Splatting~\cite{3dgs} and our approach may benefit from recent improvements to overcome its limitations.
Notably, we could integrate recent works to allow relighting of 3D Gaussians~\cite{relightable-3dgs, relightable-3dgs2} in particular when compositing multiple generated results.
While our generative backbone is memory-efficient, 3D Gaussians may exceed \SI{1}{\giga\byte} for some scenes. Recent methods could help compress them~\cite{reducing-mem-3dgs, compact-3dgs}.

\section{Conclusion}
We propose \methodname, a framework for fast, controllable, photorealistic 3D scene synthesis from a single real-world exemplar taken from in-the-wild video footage.
We employ 3D Gaussian Splatting for its speed, high-quality appearance, and natural composability.
To amortize the generation of a complex and unstructured volume of 3D Gaussians, we introduce a two-stage approach.
Each scene is first abstracted by sparse featurized voxels, which allows us to leverage Generative Cellular Automata (GCA), an efficient sparse voxel-based generative model.
We achieve single-exemplar training of GCA by injecting semantically-aware features, and employing a small network trained with random augmentations within less than 10 minutes.
In a second stage, we introduce a sparse patch-based consistency step to efficiently transform the coarse-generated voxels into a high-quality 3D Gaussian representation.
Complete generation takes 0.5 to 2 seconds for each scene, enabling truly interactive authoring sessions. 
Through various examples and a fully self-contained GUI editor, we show that our method can be used to model and synthesize various distributions and enable diverse applications: controllable scene generation, appearance transfer, mixing components from various scenes, etc.

\section*{Acknowledgement}

This work was supported by Institute of Information \& communications Technology Planning \& Evaluation (IITP) grant funded by the Korea government (MSIT) [RS-2023-00216821, Development of Beyond X-verse Core Technology for Hyper-realistic interactions by Synchronizing the Real World and Virtual Space and RS-2021-II211343, Artificial Intelligence Graduate School Program (Seoul National University)].

{\small
\bibliographystyle{cvm}
\bibliography{biblio}
}

\ifconference
\FloatBarrier
\fi

\noindent\hbox{}  %

\end{document}